\documentclass{article}

\usepackage[preprint]{neurips_2026}

\usepackage[utf8]{inputenc} 
\usepackage[T1]{fontenc}    
\usepackage{hyperref}       
\usepackage{url}            
\usepackage{booktabs}       
\usepackage{amsfonts}       
\usepackage{nicefrac}       
\usepackage{microtype}      
\usepackage{xcolor}         
\usepackage{graphicx}

\usepackage{amsmath}
\usepackage{amssymb}
\usepackage{mathtools}
\usepackage{amsthm}
\usepackage{float} 
\usepackage{bbm}
\usepackage{caption}
\usepackage{subcaption}
\usepackage{tabularx}
\usepackage{enumerate}
\usepackage{bm}
\RequirePackage{algorithm,algorithmic}
\usepackage{multirow}
\usepackage[most]{tcolorbox}
\usepackage{enumitem}
\tcbuselibrary{breakable}

\newtcolorbox{mycallout}[2][breakable, enhanced]{
  colback=blue!5!white,
  colframe=blue!75!black,
  fonttitle=\bfseries,
  coltitle=white,
  title=#2,
  enhanced,
  attach boxed title to top left={yshift=-2mm, xshift=2mm},
  boxed title style={colback=blue!75!black},
  #1
}

\title{Using Probabilistic Programs to Train \\ Inductive Reasoning in Large Language Models}

\author{%
  Liyi~Zhang \\
  Department of Computer Science\\
  Princeton University\\
  \texttt{zhang.liyi@princeton.edu} \\
  \And
  Akshay K. Jagadish \\
  Princeton AI Lab\\
  Princeton University\\
  \texttt{akshay.jagadish@princeton.edu} \\
  \And
  Brenden M.~Lake \\
Department of Psychology and Computer Science\\
Princeton University \\
\texttt{brenden@princeton.edu}
  \And
Thomas L.~Griffiths \\
Department of Psychology and Computer Science\\
Princeton University \\
\texttt{tomg@princeton.edu}
}

\begin{document}

\maketitle

\begin{abstract}
    Post-training Large Language Models (LLMs) for reasoning typically focuses on deductive tasks such as mathematics and coding where correctness is verifiable. Yet, many real-world reasoning problems are inductive: agents must infer uncertain beliefs from sparse, ambiguous observations. There are challenges to using standard fine-tuning methods for inductive reasoning, including difficulties in curating large-scale, high-quality labeled datasets and in handling targets that are inherently distributional. In this work, we introduce a novel approach, called Program-based Posterior Training (PPT), to address these limitations: we use an LLM to generate diverse open-world scenarios as probabilistic programs, run probabilistic inference to produce distributional target responses to queries, and then fine-tune on these probabilistic soft labels. Using this approach, we fine-tune LLMs on 10,000 programmatically generated scenarios and evaluate on held-out motifs, human-labeled judgments, and external benchmarks. Overall, PPT substantially improves estimation accuracy on held-out inductive tasks, increases alignment with human judgments, and transfers to external benchmarks for estimation and calibration. Additionally, the gains in raw calibration are not subsumed by post-hoc temperature scaling, showing that the models have more deeply internalized uncertainty compared to output rescaling. Together, these results suggest that probabilistic-program-mediated fine-tuning is a promising approach for post-training LLMs to reliably perform approximate inductive inference.
\end{abstract}


\section{Introduction}

Large language models (LLMs) have made rapid progress on reasoning tasks, especially in domains such as mathematics and coding \citep{WeiCoT, chen2021evaluating}. These domains largely involve deductive problems: the model is expected to derive a correct answer through a sequence of logical, deterministic intermediate steps. However, many real-world reasoning problems are inductive rather than deductive \cite{Griffiths2008, tenenbaum2011how}. People routinely infer broader patterns from limited evidence and make predictions in situations where the relevant latent factors are only partially observed \cite{Griffiths2008}. Scientific reasoning, forecasting, diagnosis, and everyday decision-making require drawing uncertain conclusions from sparse and noisy data. This requires inferring latent quantities from noisy observations and making calibrated predictions \cite{Blei2014BuildCC, bishop2007}.

Training LLMs to perform inductive inference is challenging because high-quality labeled data is difficult to obtain and the underlying problem is often inherently probabilistic, leading to answers that require uncertainty representation. Human judgments can be collected, but they are expensive and hard to scale across the diverse open-world scenarios where induction is needed. Directly asking stronger LLMs to label such scenarios is scalable \cite{jagadish2024human, jagadish2025meta}, but can inherit their biases and errors \cite{chhikara2025mind}. In addition, real-world probabilistic reasoning tasks often lack known latent variables or calibrated posterior targets. To better train LLMs to perform inductive inference, we need natural-language problems that are rich enough to resemble open-world inductive reasoning but have labels that are grounded in a well-defined probabilistic model.

\begin{figure}[t]
    \centering
    \includegraphics[width=\linewidth]{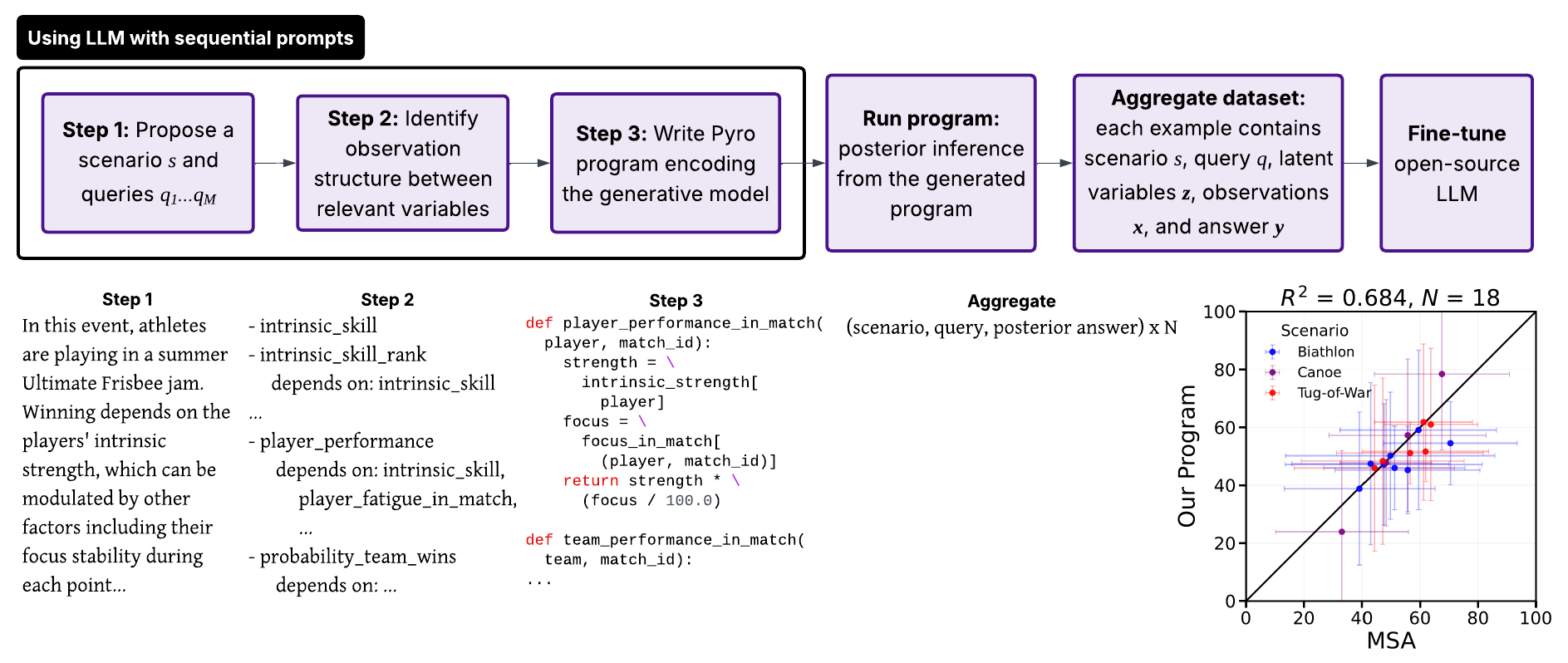}
    \caption{Workflow from data generation to LLM fine-tuning. A sequential prompting process with three steps results in probabilistic programs in Pyro for synthesized natural-language scenarios. These programs are  run to generate (scenario, query, posterior answer) tuples. These tuples are used to fine-tune LLMs. The plot in the bottom right shows that the resulting   
    pipeline programs produce aligned results with open-source programs generated by MSA on the same scenarios.}
    \label{fig:msa}
\end{figure}

Here, we address this gap by developing a method for curating large-scale natural language inductive inference data. Probabilistic programs can provide a bridge between open-world natural language and precise assumptions and supervision. LLMs are useful for proposing diverse scenarios and translating them into structured generative models \cite{rmus2026generating}, while probabilistic inference supplies training targets that are more grounded than direct LLM labels.
We introduce 
a pipeline for scalable generation of inductive reasoning datasets for fine-tuning LLMs (Figure \ref{fig:msa}). Inspired by the Model Synthesis Architecture (MSA) \cite{wong2025modeling}, our approach generates finetuning data from diverse open-world problems with a sequential prompt procedure: (1) synthesizing scenarios from a strong closed-source LLM; (2) generating probabilistic programs to capture these scenarios with an LLM; and (3) running probabilistic inference given these programs (or forward simulation) to calculate posterior distributions over relevant variables. We then use those posterior distributions as  targets for finetuning. 
With this pipeline, we create over 10,000 unique open-world scenarios with over 50,000 queries and fine-tune LLMs on the resulting data. We then evaluate the resulting models on held-out inductive tasks, compare with human judgments on sports domains, and measure accuracy and calibration on external benchmarks.

 Our approach allows us to train models on natural-language inductive reasoning problems whose latent structure and uncertainty are explicitly represented. We find that this approach improves estimation accuracy on held-out inferential structures, increases agreement with human judgments, improves performance on several transfer benchmarks, and yields better raw calibration while remaining complementary to post-hoc temperature scaling. These results suggest that data curation with probabilistic program mediation is a promising direction for training LLMs to approximate inductive inference.

In summary, we make the following contributions:
\begin{itemize}[leftmargin=0.2in,topsep=0in]
    \item We study whether large language models can be fine-tuned to improve on \emph{inductive reasoning}---inferring uncertain latent properties and predictions from sparse natural-language observations--- as opposed to deterministic, deductive reasoning.
    \item We create a dataset with over 10,000 natural-language inductive inference scenarios with probabilistically grounded supervision. To do so, we use a scalable extension of MSA that generates open-world scenarios, synthesizes probabilistic programs for these scenarios, and conducts probabilistic inference to produce targets for fine-tuning.

    \item We fine-tune LLMs on the resulting data and evaluate on held-out motifs and human judgments, showing improved estimation accuracy and stronger agreement with human judgments on unseen inferential structures.

    \item We demonstrate the utility of these principled probabilistic labels by showing that fine-tuning on a span of probabilistic labels outperforms 1) directly calling stronger, closed-source LLM; 2) fine-tuning only on the mean of the same distribution.

    \item We evaluate transfer and calibration on seven  benchmarks, showing improvements on OpenEstimate \cite{marzoev2026openestimate} and six multiple-choice benchmarks \cite{winogrande,hendrycks2021math,lin-etal-2022-truthfulqa, zellers2019hellaswag,clark2018think,bayesian-teaching}, as well as improved calibration that is complementary to post-hoc methods such as temperature scaling \cite{guo2017calibration}.
\end{itemize}

\section{Related Work}
\label{sec:related-work}

\textbf{Inductive reasoning in LLMs.}
Inductive reasoning has long been studied as probabilistic inference over structured generative models~\citep{Griffiths2008, lake-inductive}, and a growing body of work asks whether LLMs can perform such inferences. 
Wong et al. \cite{wong2023word} explore rational meaning construction, in which an LLM translates natural language into probabilistic programs in the language-of-thought tradition \cite{tenenbaum2011how, goodman2014concepts}.
Wang et al. \cite{wang2024hypothesis} formulate inductive reasoning as hypothesis search, in which an LLM proposes natural-language hypotheses and translates them into deterministic Python programs that are verified against examples. 
Yang et al. \cite{yang-etal-2024-language} evaluate LLMs as inductive reasoners over natural-language facts and rules.

\textbf{Probabilistic programs and LLMs.}
Several recent works combine LLMs with probabilistic programs at inference time.
Wu et al. \cite{wu2022foundation} train neural amortized inference networks for probabilistic programs. 
The Model Synthesis Architecture (MSA) proposed by Wong et al. \cite{wong2025modeling} addresses open-world inductive reasoning by using an LLM to synthesize a task-specific probabilistic model written in WebPPL \cite{dippl} from a natural-language scenario. 
Given sparse observations $\boldsymbol{x}$, the goal is to answer a query $\boldsymbol{y}$ by inferring latent variables $\boldsymbol{z}$. A probabilistic model specifies the distribution
\begin{align}
    p_\theta(\boldsymbol{z}, \boldsymbol{x}, \boldsymbol{y}) = p_\theta(\boldsymbol{z})\,p_\theta(\boldsymbol{x} \mid \boldsymbol{z})\,p_\theta(\boldsymbol{y} \mid \boldsymbol{z}, \boldsymbol{x}),
    \label{eq:joint}
\end{align}
and inference requires computing posterior quantities such as,
\begin{align}
    p_\theta(\boldsymbol{z} \mid \boldsymbol{x}) &\propto p_\theta(\boldsymbol{z})\,p_\theta(\boldsymbol{x} \mid \boldsymbol{z}) \\
    p_\theta(\boldsymbol{y} \mid \boldsymbol{x}) &= \int p_\theta(\boldsymbol{y} \mid \boldsymbol{z}, \boldsymbol{x})\,p_\theta(\boldsymbol{z} \mid \boldsymbol{x})\,d\boldsymbol{z}.
    \label{eq:query-posterior}
\end{align}
Given a scenario, a series of prompts leads an LLM to represent it with a probabilistic program. The LLM first identifies relevant latent factors, observations, and query variables, then writes a probabilistic program in WebPPL encoding their generative relationships. The probabilistic program is run to obtain posterior estimates over $\boldsymbol{z}$ and $\boldsymbol{y}$. Thus, the LLM supplies open-world knowledge and the probabilistic program formalizes the precise meaning, facilitating coherent posterior inference.


 \textbf{Fine-tuning LLMs with probabilistic models.} A line of work suggests that autoregressive modeling can support implicit Bayesian inference \cite{xie2022an, zhang2025what}, motivating the use of probabilistic models as training signals for LLMs. Recent work has begun to explore this direction. Hu et al. \cite{hu2024amortizing} fine-tune LLMs via GFlowNets to sample from intractable posteriors over latent chains of thought. 
Hollman et al. \cite{hollman2025nature} use probabilistic models to train transformers to fill in missing values in tabular data. Qiu et al. \cite{bayesian-teaching} fine-tune LLMs to mimic a hand-coded Bayesian assistant in a flight-recommendation task and demonstrate transfer to structurally similar recommendation tasks. 
Their 
supervision target is the assistant's point recommendation at each round.  Our approach extends this significantly: it synthesizes a distinct probabilistic program for each open-world scenario, uses the full posterior as a soft label, and evaluates transfer across heterogeneous inferential motifs and external benchmarks. 

 \textbf{Uncertainty calibration in LLMs.}
Calibration of LLM outputs has been studied through both prompting and post-hoc adjustment. Chhikara et al. \cite{chhikara2025mind} document persistent overconfidence and distractor effects in current LLMs. Tiang et al. \cite{tian2023just} establish that models can be prompted to produce well-calibrated verbalized confidences; Shen et al. \cite{pmlr-v235-shen24c} propose Thermometer, a post-hoc method that improves calibration across tasks. Post-hoc temperature scaling~\citep{guo2017calibration} remains the standard reference baseline. These methods supervise or adjust meta-confidence over an existing answer distribution. 
We take a different approach: rather than tuning meta-confidence, we supervise the answer distribution with the Bayesian posterior. The resulting calibration improvements compose with temperature scaling, 
which suggests that it shifts the model's raw uncertainty representation rather than rescaling its outputs. 


\section{Program-based Posterior Training}
\label{sec:method}

\begin{figure}
    \centering
    \begin{subfigure}{0.36\textwidth}
        \includegraphics[width=\linewidth]{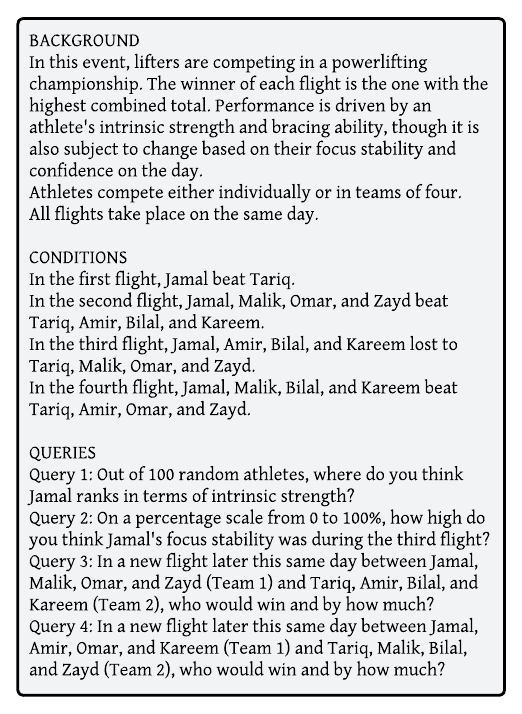}
        \captionsetup{skip=5pt}
        \caption{Scenario and query text.}
    \end{subfigure}
    \begin{subfigure}{0.63\textwidth}
        \includegraphics[width=\linewidth]{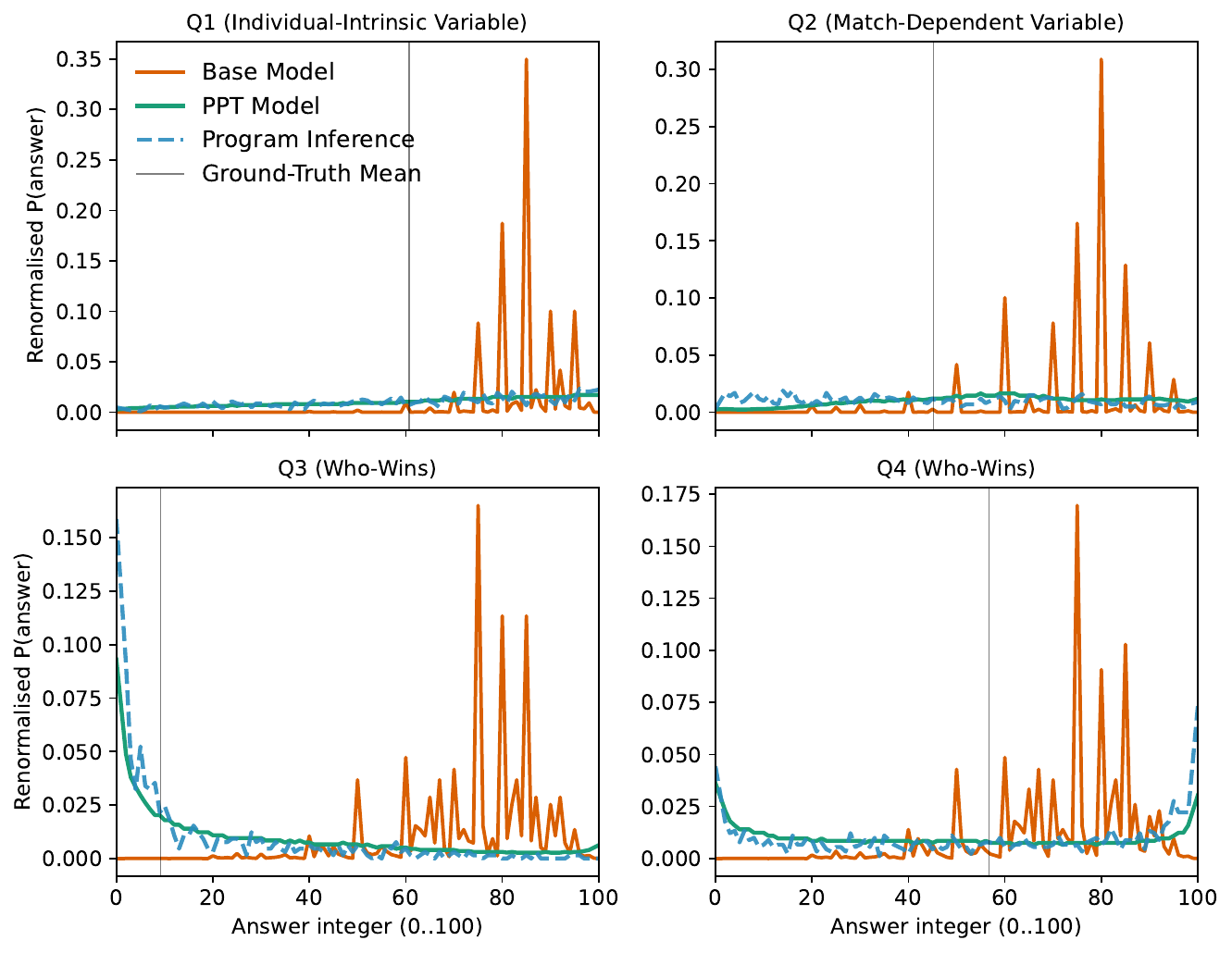}
        \caption{LLM posteriors and program-inferred posteriors (ground truth).}
    \end{subfigure}
    \caption{Effects of Program-based Posterior Training. (a) Example held-out scenario and queries. (b) Posteriors over answer-token probabilities from the base LLM and LLM fine-tuned on data from probabilistic programs (\textsc{Llama3-8B-Instruct}). Blue dashed line is programmatic inference result, treated as ground truth. A text appended before the scenario (see Appendix Section \ref{sec:appendix-imp}) instructs the LLM that Q3 and Q4 are estimated on 0-100 scale, where lower number means Team 1 more likely wins. LLM posteriors are more similar to those of the program.}
    \label{fig:example-posterior}
\end{figure}

Our goal is to construct natural-language inductive inference data with probabilistically grounded supervision, and then use this data to fine-tune LLMs. To do so, we propose the Program-based Posterior Training (PPT) pipeline, illustrated by Figure \ref{fig:msa}. The data-generation stage proceeds in three steps. First, it uses an LLM to generate diverse natural-language scenarios and queries. Second, an LLM is called to translate each scenario into a probabilistic program in Pyro \cite{bingham2019pyro}. Third, probabilistic inference or forward simulation is run in the resulting program to produce supervised training targets. We then fine-tune open-source LLMs on these scenario--query--answer examples.

\subsection{Data generation}

Our data-generation pipeline 
(Figure \ref{fig:msa}) builds on MSA \cite{wong2025modeling}, although it uses MSA as a scalable data curation mechanism rather than as an inference-time cognitive model. In the original MSA setting, human-written scenarios are translated into WebPPL programs and evaluated against human judgments. 
We extend this procedure in two ways. First, the LLM is also used to generate the natural-language scenarios themselves, allowing the pipeline to produce many scenario--query pairs across domains. Second, we translate the synthesized probabilistic programs into Pyro, which allows us to use Python-based probabilistic programming and PyTorch infrastructure. Modern LLMs are generally stronger at writing and editing Python code \cite{twist2025llms} than WebPPL code, which improves the reliability and expressivity of program synthesis and debugging.

Each example consists of a natural-language scenario $s$, a query $q$, latent variables $\boldsymbol{z}$, observations $\boldsymbol{x}$, and an answer variable $\boldsymbol{y}$. The scenario describes an open-world inductive setting with observed and latent properties (which are probed in the queries). Figure \ref{fig:example-posterior} shows an example scenario and inferred posteriors on the queries. The LLM is sequentially prompted with in-context examples to:
\begin{enumerate}
    \item Propose a scenario $s$ and queries $q_1,...,q_M$,
    \item Identify the relevant latent variables $\boldsymbol{z}$ and observation structure $p_\theta(\boldsymbol{z}, \boldsymbol{x}, \boldsymbol{y})$,
    \item Write a Pyro program encoding the corresponding generative model.
\end{enumerate}
We use the same framing of probabilistic quantities as MSA, where the program defines the joint distribution $p_\theta(\boldsymbol{z}, \boldsymbol{x}, \boldsymbol{y})$ from Equation \ref{eq:joint}.
To answer the queries in scenarios, the program is run to conduct posterior inference using MCMC or Rejection Sampling, giving the posterior of the quantity being queried given observations, $p_\theta(\boldsymbol{y} | \boldsymbol{x})$ from Equation \ref{eq:query-posterior}.



As a sanity check, we compared the 
Pyro programs we generated with the publicly released MSA programs on matched scenarios and queries. We found that the posterior estimates were in strong alignment across scenarios (Figure \ref{fig:msa} bottom right).

Table \ref{tab:dataset-coverage} counts scenarios and queries with posterior answers by domain. The domains we choose are \textit{Sports} (that MSA also uses), \textit{Healthcare}, \textit{General}. These domains use the identical pipeline, except each domain uses a separate prompt in the Step 1 scenario generation stage. They also share the structure in terms of background, conditions (between 3 and 4), and queries on an entity-intrinsic property, an observation-episodic variable, and two group comparisons. We provide specific scenario examples in Appendix Section \ref{sec:appendix-imp}.

\begin{table}[t]      
  \centering  
  \caption{Dataset coverage: count of scenarios and queries.} 
  \resizebox{0.45\textwidth}{!}
  {
  \begin{tabular}{lcccc}
  \toprule                       
   & Total & Sports & Healthcare & General \\
  \midrule                         
  Scenarios & $14{,}912$ & $4{,}866$ & $5{,}013$ & $5{,}033$ \\
  Queries   & $59{,}633$ & $19{,}451$ & $20{,}050$ & $20{,}132$ \\   
  \bottomrule                         
  \end{tabular}}                          
  \label{tab:dataset-coverage}  
  \end{table} 

\subsection{Fine-Tuning Format}

\textbf{Posterior sampling.} We use two types of supervision from the synthesized programs. The first type is based on  \textit{posterior inference}. The posterior $p_\theta(\boldsymbol{y} | \boldsymbol{x})$ is acquired by running the Pyro program described above. It then becomes training labels for the answer to the query in the scenario in one of two ways: 1) a discretized posterior distribution; 2) a single representative answer, which is the posterior mean. In 1), the discretized posterior is realized as probabilities on answer-tokens and treated as the fine-tuning target for the LLM cross-entropy loss. 

Specifically, let the posterior $p_\theta(\boldsymbol{y} | \boldsymbol{x})$ have a finite domain, and let $\{b_0,...,b_{K}\}$ be boundaries of this domain evenly partitioned into $K$ bins. We define $p_k$ as the posterior probability over bin $k$: $p_k = \int_{b_{k}}^{b_{k+1}}p_\theta(\boldsymbol{y} | \boldsymbol{x}) d\boldsymbol{y}$. Then, the LLM cross-entropy loss is:
\begin{align*}
    \mathcal{L_{\text{Dist}}} = -\sum_{k=0}^{K-1} p_k \log \hat{p}_k,
\end{align*}
where $\hat{p}_k$ is the LLM-predicted probability for ``the token of $p_k$''. In practice, we scale dataset query answers to the 0 to 100 range, and $p_k$ represents the density around an integer, say 36. Then ``the token of $p_k$'' refers to the token for the Arabic number 36.

We also explore point-target fine-tuning. This second form trains on the posterior mean with probability $1$, so the LLM essentially performs regular next-token prediction for the posterior mean rounded to the nearest integer.

\textbf{Forward sampling.} As an alternative to posterior sampling, a supervision can also use \textit{forward sampling}. Instead of conditioning on observations and running posterior inference, we sample latent variables and outcomes directly from the generative model. Forward sampling is noisier because each sampled example reflects one draw from the model rather than an inferred posterior. However, it is cheaper: it avoids iterative posterior inference and can generate additional scenario--query--answer pairs with fewer inference-time costs, leading to hundreds of times more datapoints generated. We therefore treat forward-sampled data as an alternative supervision source.

\subsection{Data splits and held-out motifs}

To evaluate generalization beyond repeated surface forms, we split data by \textit{motifs} rather than only by individual examples. A motif denotes an underlying inferential structure, implemented as patterns in the observation process. For example, one motif is ``one person consistently wins but loses one match''. We replicate the motifs from \textsc{MSA} \citep{wong2025modeling}. Training and validation examples may share broad domains, but held-out motif evaluation requires the model to generalize to new latent structures. This is stricter than a random split over paraphrases or entities, because the model cannot solve the evaluation set by memorizing a fixed scenario template.





\section{Empirical Evaluations}
\label{sec:exp}

We evaluate whether PPT improves inductive reasoning in LLMs by testing agreement with Pyro posteriors on held-out scenarios, comparisons with strong base LLMs on these scenarios, agreement with human judgments, and transfer/calibration on external benchmarks.

\textbf{Implementational details} We use low-rank adaptation (LoRA; \cite{hu2022lora}) to fine-tune \textsc{Llama-3-8B-Instruct} and \textsc{Qwen-2-7B-Instruct} \cite{grattafiori2024llama3herdmodels, yang2024qwen2technicalreport}. For more details, see Appendix Section \ref{sec:appendix-imp}.\footnote{Code is available at \url{https://github.com/zhang-liyi/llm-inductive}.}

\begin{figure*}[t]
    \centering
    \includegraphics[width=0.8\textwidth]{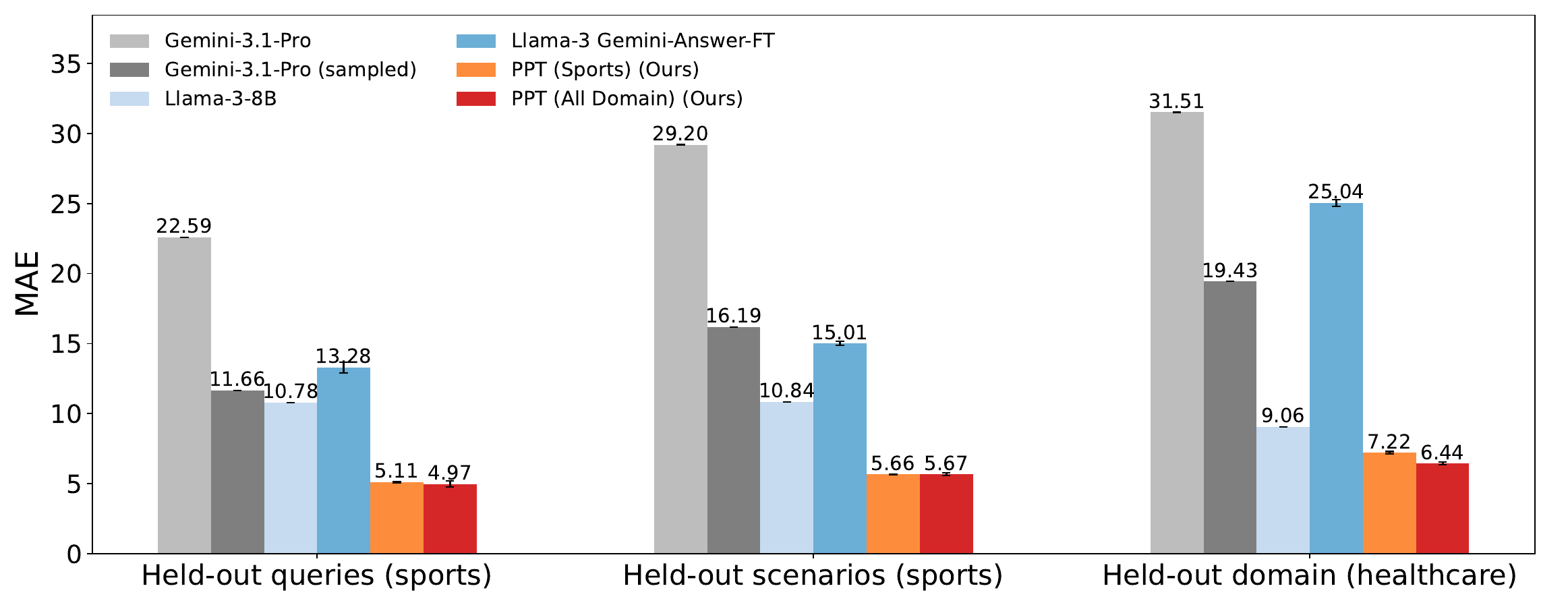}
    \caption{Mean absolute error (MAE) where colors indicate different models, base and fine-tuned. Axis markers indicate the evaluation data: Held-out queries uses sports training scenarios, held-out scenarios have unseen motifs, and held-out domain (healthcare) is where PPT (Sports) has not trained on but PPT (All Domain) has seen. For open-source models, we take the mean response across integers 0 to 100. For closed-source models, we use greedy sampling (\textsc{Gemini-3.1-Pro}) or average across 100 random seeds with temperature = 1 (\textsc{Gemini-3.1-Pro} sampled). As another baseline, \textsc{Llama-3}-\textsc{Gemini}-Answer-FT is \textsc{Llama-3} trained on targets directly generated by \textsc{Gemini} rather than by using probabilistic programs. {While seen scenarios are easiest for the our fine-tuned models, these models also generalize to new motifs and new domains, significantly outperforming the base LLM. Diversifying training domains gives a further improvement on LLM performance (All-Domain vs Sports).}}
    \label{fig:mae-bar}
\end{figure*}

\subsection{Evaluating fine-tuned LLMs against MSA and human labels}

\subsubsection{In-domain: sports scenarios}

We first evaluate models on sports-domain scenarios similar to those that were used to evaluate MSA. These scenarios describe novel sports competitions with sparse observations about athletes or teams and query latent properties or future outcomes. Our validation split uses held-out scenarios with unseen motifs, so evaluation requires generalization to new inferential structures rather than memorization of templates.

\textbf{Fine-tuned models represent uncertainty structures} Figure~\ref{fig:example-posterior} shows a representative held-out scenario. We compare the answer distributions produced by the base LLM, the LLM fine-tuned by PPT, and the Pyro program. The base model produces uncalibrated responses, often placing probability mass in uneven and idiosyncratic ways (e.g., on multiples of ten; \cite{mccoy2023embers}), while the fine-tuned model produces a posterior closer to the distributions computed with probabilistic inference. 

We further verify whether the LLM constructs a consistent model behind scenarios by checking several similar scenarios with known progressively increasing answers on one query. We manually edit one scenario by performing one action at a time (flip a win or loss or to swap two players), creating 8 scenarios labeled S1-S8, where a player named Jamal progressively goes from weaker to stronger. We found that all methods correctly show a progressive increase in its mean estimate of Jamal's strength. However, the base LLM plateaus more than the PPT-LLM, which continues to increase by 1-2\% each step for scenarios S7 and S8 and, as a result, maintains closer estimates with the program; see Figure \ref{fig:progressive} in the Appendix Section \ref{sec:appendix-exp}.

Next, we consider the overall quantitative comparison on large-scale held-out scenarios as shown in Figure~\ref{fig:mae-bar}. The Figure reports mean absolute error (MAE) on three evaluation sets: sports training scenarios with held-out queries, sports held-out scenarios with unseen motifs, and healthcare scenarios from held-out domain.  
We find that the LLM fine-tuned with PPT (red and orange) has substantially lower mean-absolute error on its posterior mean compared to the base LLM. These results illustrate the intended effect of our approach: the model does not merely shift its point prediction, but learns to assign probability mass in a way that better reflects the uncertainty structure of the underlying probabilistic model.

\textbf{Strong closed-source LLM also fails to answer probabilistic queries well} These problems are also challenging to strong closed-source models like \textsc{Gemini-3.1-Pro} (Figure~\ref{fig:mae-bar}), which, surprisingly, gives worse performance than \textsc{Llama-3-8B}. Each model's mean response over the possible answer domain is more accurate than its argmax response. For \textsc{Gemini}, its mean response is approximated by averaging over 100 samples.

\begin{figure*}[t]
    \centering
    \begin{subfigure}{0.49\textwidth}
        \centering
        \includegraphics[width=\textwidth]{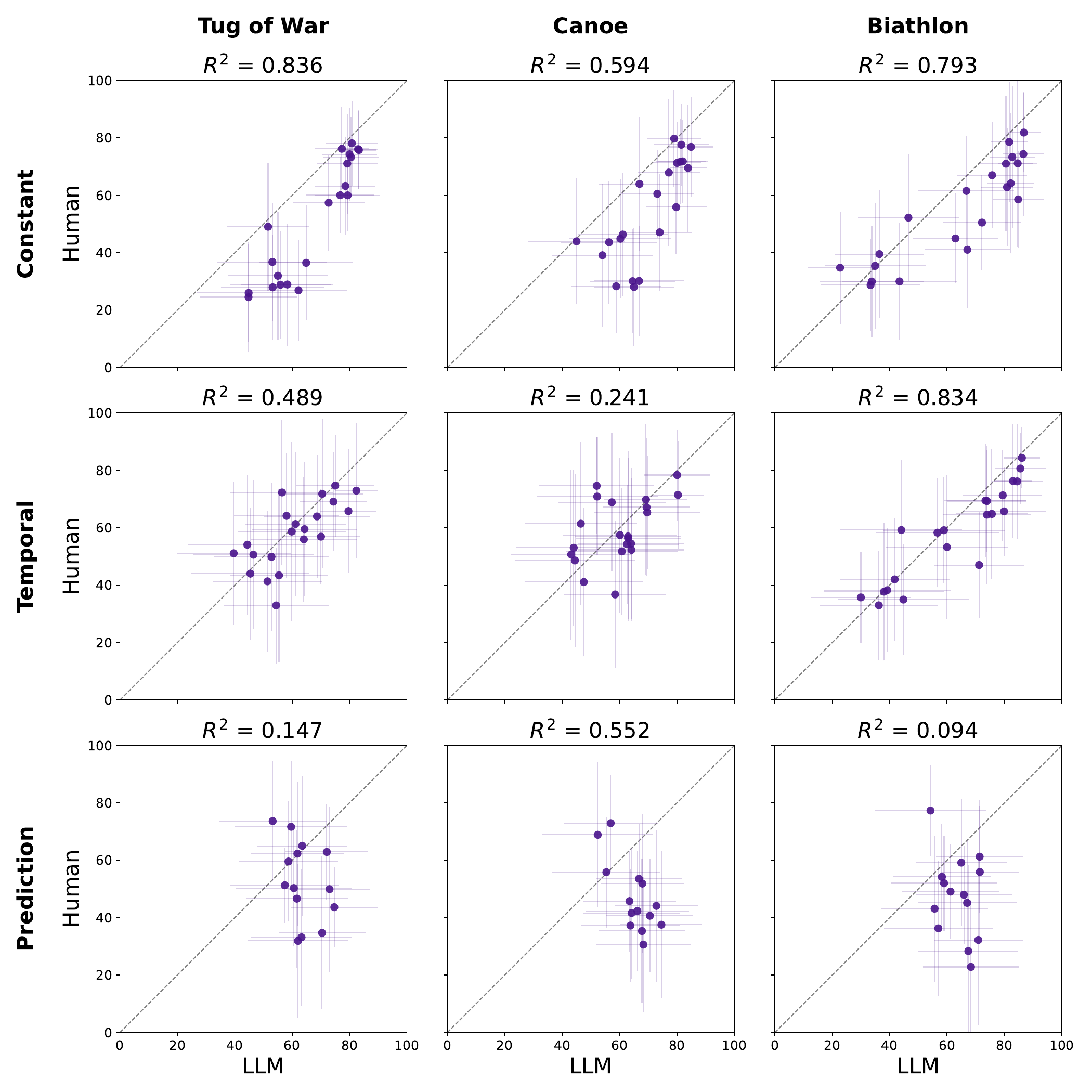}
        \caption{Base pretrained LLM.}
    \end{subfigure}
    \begin{subfigure}{0.49\textwidth}
        \centering
        \includegraphics[width=\textwidth]{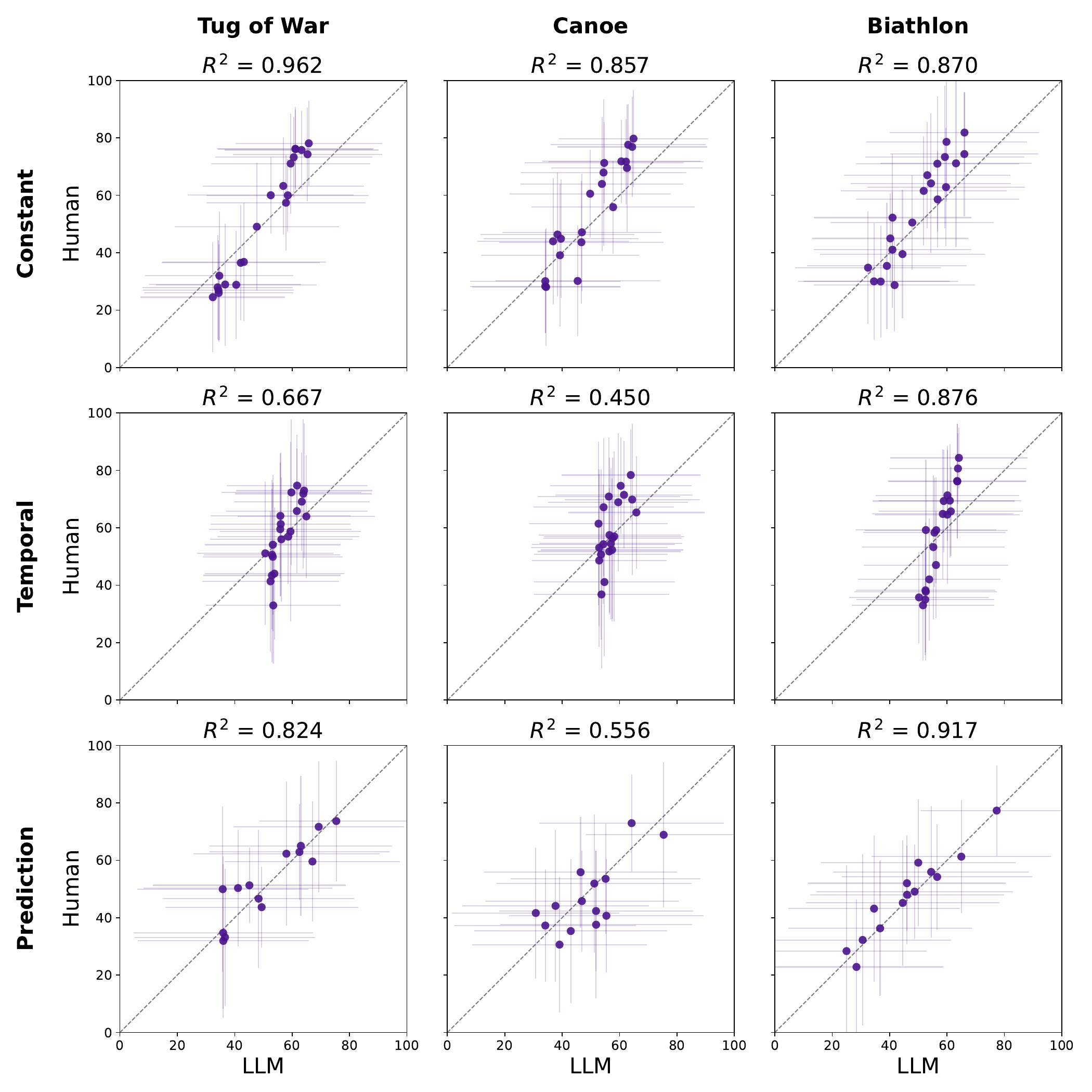}
        \caption{Fine-tuned LLM.}
    \end{subfigure}
    \caption{Per-query mean response and standard-deviation between humans and LLMs on held-out sports scenarios. {Fine-tuned LLMs align much more closely with human responses.} The overall correlation $R^2$ with humans for base LLM, fine-tuned LLM, and MSA results from \cite{wong2025modeling} are: 0.509, 0.775, 0.769, suggesting that our approach achieves similar alignment to probabilistic programs. }
    \label{fig:in-domain-grids}
\end{figure*}

\textbf{Fine-tuned models match human results more closely} We compare LLM predictions with human judgments. For each scenario and query, we aggregate model responses and human responses into mean estimates with standard deviation. Figure~\ref{fig:in-domain-grids} shows that our approach substantially improves alignment with human judgments across held-out sports scenarios. After fine-tuning, the model's responses move closer to the diagonal across most query types, with lower error and stronger correlation. On an aggregate level, we find that LLM fine-tuned with PPT ($0.775$) had the highest overall correlation ($R^2$) with humans compared to base LLM (0.509) and MSA-generated program (0.769), respectively. This suggests that our approach improves  agreement with both the true posterior and human judgments.

\subsubsection{Out-of-domain: healthcare scenarios}

We then test whether the gains from our approach transfer beyond the domain used for fine-tuning. We evaluate on healthcare-domain scenarios generated by the same pipeline, where the model must infer latent patient- or treatment-relevant quantities from sparse observations (for an example, see Appendix Section \ref{sec:appendix-imp}). This provides an out-of-domain test for the sports-fine-tuned model, since the surface domain, entities, and scenario semantics differ from the sports training data.

\textbf{Fine-tuning reduces MAE} Figure~\ref{fig:mae-bar} reports mean absolute error (MAE) on in-domain and out-of-domain evaluation sets. Sports-only and all-domain fine-tuning using generated targets generated using probabilistic programs substantially reduce MAE on all sets. 

The healthcare validation set provides a stronger test of domain transfer. The sports-fine-tuned model improves over the base model despite not being trained on healthcare scenarios, suggesting that our approach teaches some domain-general inductive behavior. 

\begin{table}[t]
\centering
\caption{Transfer to external benchmarks on \textsc{Llama3-8B} and \textsc{Qwen2-7B}.
NLL/ECE include a temperature-scaling (TS) block.
Best per column in \textbf{bold} (within each block).
On the 3-class BT datasets, TS lands in a high-$T$ regime where the
predictive distribution approaches uniform (NLL\,$\to\ln 3\approx 1.099$):
Base+TS hits the optimizer's $T{=}100$ ceiling,
while \textsc{PPT}+TS finds an interior optimum at $T\approx 4$--$7$ that retains
only marginal signal over uniform.}
\label{tab:calibration}

\subcaption{NLL across benchmarks (lower is better).}
\label{tab:nll}
\resizebox{\textwidth}{!}{%
\begin{tabular}{lllllllll}
\toprule
Method: \textsc{Llama-3} & OE & BT & BT-guided & MMLU & TruthfulQA & HellaSwag & ARC-C & Winogrande \\
\midrule
Base            & $7.09$        & $2.91$        & $2.12$        & $1.69$        & $3.49$        & $1.26$        & $0.98$        & $1.15$ \\
\textsc{PPT} (Distribution)     & \underline{\bm{$4.53$}}
                  \scriptsize{$\pm 0.02$}
                & \underline{\bm{$1.25$}}
                  \scriptsize{$\pm 0.01$}
                & \underline{\bm{$1.13$}}
                  \scriptsize{$\pm 0.00$}
                & \underline{\bm{$1.01$}}
                  \scriptsize{$\pm 0.01$}
                & \underline{\bm{$1.78$}}
                  \scriptsize{$\pm 0.02$}
                & \underline{\bm{$0.88$}}
                  \scriptsize{$\pm 0.02$}
                & \underline{\bm{$0.61$}}
                  \scriptsize{$\pm 0.01$}
                & \underline{\bm{$0.80$}}
                  \scriptsize{$\pm 0.01$} \\
\textsc{PPT} (Mean) & $5.08$
                  \scriptsize{$\pm 0.02$}                 & $1.86$
                  \scriptsize{$\pm 0.00$}                 & $1.37$
                  \scriptsize{$\pm 0.00$}                 & $1.27$
                  \scriptsize{$\pm 0.01$}                 & $2.36$
                  \scriptsize{$\pm 0.02$}                 & $0.98$
                  \scriptsize{$\pm 0.01$}                 & $0.74$
                  \scriptsize{$\pm 0.00$}                 & $0.91$
                  \scriptsize{$\pm 0.01$} \\
\midrule
Base + TS           & ---            & ---            & ---            & $1.01$        & $1.76$        & $0.95$        & $0.62$        & $0.84$ \\
\textsc{PPT} (Distribution) + TS    & ---
                & \underline{\bm{$1.08$}}
                  \scriptsize{$\pm 0.00$}
                & \underline{\bm{$1.09$}}
                  \scriptsize{$\pm 0.00$}
                & \underline{\bm{$0.96$}}
                  \scriptsize{$\pm 0.01$}
                & \underline{\bm{$1.62$}}
                  \scriptsize{$\pm 0.03$}
                & \underline{\bm{$0.87$}}
                  \scriptsize{$\pm 0.02$}
                & \underline{\bm{$0.60$}}
                  \scriptsize{$\pm 0.01$}
                & \underline{\bm{$0.80$}}
                  \scriptsize{$\pm 0.00$} \\
\midrule\midrule
Method: \textsc{Qwen-2} & OE & BT & BT-guided & MMLU & TruthfulQA & HellaSwag & ARC-C & Winogrande \\
\midrule
Base            & $6.91$ & $2.56$ & $2.21$ & $1.96$ & $4.81$ & $0.84$ & $0.90$ & $2.18$ \\
\textsc{PPT} (Distribution)     & \underline{\bm{$3.86$}}
                  \scriptsize{$\pm 0.00$}                 & \underline{\bm{$1.40$}}
                  \scriptsize{$\pm 0.01$}                 & \underline{\bm{$1.28$}}
                  \scriptsize{$\pm 0.01$}                 & \underline{\bm{$1.12$}}
                  \scriptsize{$\pm 0.01$}                 & \underline{\bm{$3.04$}}
                  \scriptsize{$\pm 0.03$}                 & \underline{\bm{$0.60$}}
                  \scriptsize{$\pm 0.00$}                 & \underline{\bm{$0.51$}}
                  \scriptsize{$\pm 0.01$}                 & \underline{\bm{$1.23$}}
                  \scriptsize{$\pm 0.01$} \\
\textsc{PPT} (Mean) & $4.59$
                  \scriptsize{$\pm 0.08$}                 & $2.06$
                  \scriptsize{$\pm 0.02$}                 & $1.77$
                  \scriptsize{$\pm 0.02$}                 & $1.67$
                  \scriptsize{$\pm 0.01$}                 & $4.48$
                  \scriptsize{$\pm 0.06$}                 & $0.75$
                  \scriptsize{$\pm 0.01$}                 & $0.76$
                  \scriptsize{$\pm 0.01$}                 & $1.88$
                  \scriptsize{$\pm 0.02$} \\
\midrule
Base + TS           & ---            & ---            & ---            & $1.01$        & $4.81$        & \underline{\bm{$0.58$}} & $0.49$        & $1.03$ \\
\textsc{PPT} (Distribution) + TS    & ---
                & \underline{\bm{$1.09$}}
                  \scriptsize{$\pm 0.00$}
                & \underline{\bm{$1.09$}}
                  \scriptsize{$\pm 0.00$}
                & \underline{\bm{$0.94$}}
                  \scriptsize{$\pm 0.01$}
                & \underline{\bm{$3.16$}}
                  \scriptsize{$\pm 0.04$}
                & $0.60$
                  \scriptsize{$\pm 0.01$}
                & \underline{\bm{$0.47$}}
                  \scriptsize{$\pm 0.01$}
                & \underline{\bm{$0.96$}}
                  \scriptsize{$\pm 0.01$} \\
\bottomrule
\end{tabular}%
}

\vspace{1em}

\subcaption{Expected Calibration Error (15-bin) (lower is better).}
\label{tab:ece}
\resizebox{\textwidth}{!}{%
\begin{tabular}{llllllll}
\toprule
Method: \textsc{Llama-3} & BT & BT-guided & MMLU & TruthfulQA & HellaSwag & ARC-C & Winogrande \\
\midrule
Base            & $0.467$        & $0.409$        & $0.281$        & $0.462$        & $0.196$        & $0.158$        & $0.289$ \\
Verbalized      & $0.255$        & $0.312$        & \underline{\bm{$0.084$}} & $0.470$        & $0.141$        & $0.211$        & $0.284$ \\
\textsc{PPT} (Distribution)     & \underline{\bm{$0.218$}}
                  \scriptsize{$\pm 0.002$}
                & \underline{\bm{$0.134$}}
                  \scriptsize{$\pm 0.007$}
                & $0.112$
                  \scriptsize{$\pm 0.006$}
                & \underline{\bm{$0.275$}}
                  \scriptsize{$\pm 0.009$}
                & \underline{\bm{$0.064$}}
                  \scriptsize{$\pm 0.010$}
                & \underline{\bm{$0.058$}}
                  \scriptsize{$\pm 0.010$}
                & \underline{\bm{$0.094$}}
                  \scriptsize{$\pm 0.009$} \\
\textsc{PPT} (Mean) & $0.380$
                  \scriptsize{$\pm 0.000$}                 & $0.246$
                  \scriptsize{$\pm 0.003$}                 & $0.216$
                  \scriptsize{$\pm 0.002$}                 & $0.377$
                  \scriptsize{$\pm 0.002$}                 & \underline{\bm{$0.060$}}
                  \scriptsize{$\pm 0.003$}                 & $0.118$
                  \scriptsize{$\pm 0.005$}                 & $0.221$
                  \scriptsize{$\pm 0.007$} \\
\midrule
Base + TS           & ---            & ---                        & $0.100$        & $0.245$        & \underline{\bm{$0.023$}}   & $0.062$        & $0.126$ \\
\textsc{PPT} (Distribution) + TS    & \underline{\bm{$0.025$}}
                  \scriptsize{$\pm 0.002$}
                & \underline{\bm{$0.020$}}
                  \scriptsize{$\pm 0.003$}
                & \underline{\bm{$0.069$}}
                  \scriptsize{$\pm 0.004$}
                & \underline{\bm{$0.223$}}
                  \scriptsize{$\pm 0.014$}
                & $0.032$
                  \scriptsize{$\pm 0.003$}
                & \underline{\bm{$0.051$}}
                  \scriptsize{$\pm 0.010$}
                & \underline{\bm{$0.080$}}
                  \scriptsize{$\pm 0.002$} \\
\midrule\midrule
Method: \textsc{Qwen-2} & BT & BT-guided & MMLU & TruthfulQA & HellaSwag & ARC-C & Winogrande \\
\midrule
Base            & $0.464$ & $0.445$ & $0.273$ & $0.475$ & $0.134$ & $0.135$ & $0.337$ \\
\textsc{PPT} (Distribution)     & \underline{\bm{$0.278$}}
                  \scriptsize{$\pm 0.005$}                 & \underline{\bm{$0.228$}}
                  \scriptsize{$\pm 0.005$}                 & \underline{\bm{$0.180$}}
                  \scriptsize{$\pm 0.001$}                 & \underline{\bm{$0.381$}}
                  \scriptsize{$\pm 0.002$}                 & \underline{\bm{$0.060$}}
                  \scriptsize{$\pm 0.001$}                 & \underline{\bm{$0.092$}}
                  \scriptsize{$\pm 0.001$}                 & \underline{\bm{$0.246$}}
                  \scriptsize{$\pm 0.003$} \\
\textsc{PPT} (Mean) & $0.415$
                  \scriptsize{$\pm 0.002$}                 & $0.386$
                  \scriptsize{$\pm 0.005$}                 & $0.254$
                  \scriptsize{$\pm 0.001$}                 & $0.475$
                  \scriptsize{$\pm 0.005$}                 & $0.119$
                  \scriptsize{$\pm 0.002$}                 & $0.128$
                  \scriptsize{$\pm 0.004$}                 & $0.324$
                  \scriptsize{$\pm 0.001$} \\
\midrule
Base + TS           & ---            & ---            & $0.133$        & $0.475$        & $0.035$        & $0.062$        & $0.128$ \\
\textsc{PPT} (Distribution) + TS    & \underline{\bm{$0.056$}}
                  \scriptsize{$\pm 0.004$}
                & \underline{\bm{$0.051$}}
                  \scriptsize{$\pm 0.006$}
                & \underline{\bm{$0.095$}}
                  \scriptsize{$\pm 0.000$}
                & \underline{\bm{$0.383$}}
                  \scriptsize{$\pm 0.003$}
                & \underline{\bm{$0.015$}}
                  \scriptsize{$\pm 0.002$}
                & \underline{\bm{$0.050$}}
                  \scriptsize{$\pm 0.006$}
                & \underline{\bm{$0.110$}}
                  \scriptsize{$\pm 0.001$} \\
\bottomrule
\end{tabular}%
}

\subcaption{Accuracy across benchmarks; OpenEstimate (OE) reported as MAE (lower is better).}
\label{tab:accuracy}
\resizebox{\textwidth}{!}{%
\begin{tabular}{lllllllll}
\toprule
Method: \textsc{Llama-3} & OE (MAE) $\downarrow$ & BT & BT-guided & MMLU & TruthfulQA & HellaSwag & ARC-C & Winogrande \\
\midrule
Base            & $29.0$         & $37.8\%$        & $35.7\%$        & $60.0\%$        & $42.2\%$        & $63.6\%$        & $78.6\%$        & $54.5\%$ \\
Verbalized      & ---            & $35.3\%$        & $35.7\%$        & $42.8\%$        & $16.0\%$        & $43.1\%$        & $54.8\%$        & $50.7\%$ \\
\textsc{PPT} (Distribution)     & \underline{\bm{$23.3$}}
                  \scriptsize{$\pm 0.3$}
                & \underline{\bm{$39.3\%$}}
                  \scriptsize{$\pm 0.0\%$}
                & \underline{\bm{$37.3\%$}}
                  \scriptsize{$\pm 0.2\%$}
                & \underline{\bm{$61.2\%$}}
                  \scriptsize{$\pm 0.3\%$}
                & \underline{\bm{$43.2\%$}}
                  \scriptsize{$\pm 0.7\%$}
                & \underline{\bm{$66.8\%$}}
                  \scriptsize{$\pm 1.2\%$}
                & \underline{\bm{$79.6\%$}}
                  \scriptsize{$\pm 0.5\%$}
                & \underline{\bm{$58.1\%$}}
                  \scriptsize{$\pm 0.4\%$} \\
\textsc{PPT} (Mean) & $23.8$
                  \scriptsize{$\pm 0.1$}                 & $36.8\%$
                  \scriptsize{$\pm 0.0\%$}                 & \underline{\bm{$37.2\%$}}
                  \scriptsize{$\pm 0.1\%$}                 & $59.5\%$
                  \scriptsize{$\pm 0.2\%$}                 & $41.7\%$
                  \scriptsize{$\pm 0.1\%$}                 & $61.6\%$
                  \scriptsize{$\pm 0.4\%$}                 & $77.4\%$
                  \scriptsize{$\pm 0.2\%$}                 & $54.2\%$
                  \scriptsize{$\pm 0.2\%$} \\
\midrule\midrule
Method: \textsc{Qwen-2} & OE (MAE) $\downarrow$ & BT & BT-guided & MMLU & TruthfulQA & HellaSwag & ARC-C & Winogrande \\
\midrule
Base            & $14.2$ & \underline{\bm{$39.6\%$}} & $37.6\%$ & $65.1\%$ & \underline{\bm{$41.1\%$}} & \underline{\bm{$80.1\%$}} & $83.6\%$ & $61.3\%$ \\
\textsc{PPT} (Distribution)     & \underline{\bm{$13.4$}}
                  \scriptsize{$\pm 0.1$}                 & $38.9\%$
                  \scriptsize{$\pm 0.3\%$}                 & \underline{\bm{$38.4\%$}}
                  \scriptsize{$\pm 0.5\%$}                 & \underline{\bm{$65.5\%$}}
                  \scriptsize{$\pm 0.1\%$}                 & $40.4\%$
                  \scriptsize{$\pm 0.1\%$}                 & $79.4\%$
                  \scriptsize{$\pm 0.1\%$}                 & \underline{\bm{$84.1\%$}}
                  \scriptsize{$\pm 0.1\%$}                 & \underline{\bm{$63.8\%$}}
                  \scriptsize{$\pm 0.4\%$} \\
\textsc{PPT} (Mean) & $14.7$
                  \scriptsize{$\pm 0.2$}                 & \underline{\bm{$39.4\%$}}
                  \scriptsize{$\pm 0.2\%$}                 & $36.9\%$
                  \scriptsize{$\pm 0.3\%$}                 & $64.9\%$
                  \scriptsize{$\pm 0.2\%$}                 & $38.8\%$
                  \scriptsize{$\pm 0.5\%$}                 & $79.3\%$
                  \scriptsize{$\pm 0.2\%$}                 & $83.7\%$
                  \scriptsize{$\pm 0.2\%$}                 & $61.1\%$
                  \scriptsize{$\pm 0.1\%$} \\
\bottomrule
\end{tabular}%
}

\vspace{1em}
\end{table}

\subsection{Fine-tuned LLMs transfer to common external benchmarks}

We finally evaluate whether training on data generated from probabilistic programs results in generalizable inductive and probabilistic reasoning capability. To do this, we consider seven different benchmarks. We consider OpenEstimate (OE; \cite{marzoev2026openestimate}) as it measures LLM's prior elicitation. We evaluate on the Bayesian Teaching dataset (BT; \cite{bayesian-teaching}) since it features multiple-choice problems that involve uncertainty and an underlying optimal Bayesian model. Finally, for a wider performance and calibration evaluation, we use MMLU \cite{hendrycks2021math}, TruthfulQA \cite{lin-etal-2022-truthfulqa}, HellaSwag \cite{zellers2019hellaswag}, ARC-Challenge \cite{clark2018think}, and Winogrande \cite{winogrande}, which involve challenging multiple-choice problems that are not purely deterministic. 



\textbf{Metrics} We use accuracy or mean-absolute error (MAE) for a simple and intuitive performance evaluation, negative log-likelihood (NLL) for how well the LLM models the ground truth answer, and expected calibration error (ECE) to measure model calibration. 

\textbf{Baselines} We use Base-Instruct LLM, Verbalized LLM \cite{tian2023just}, and temperature scaling (TS; \cite{guo2017calibration}). Verbalized LLM is the base LLM that explicitly outputs confidence level on multiple choices in its response. This baseline tests whether the base LLM already exhibits reliable and extractable confidence levels beyond token-level probabilities. TS is a post-hoc calibration method that tunes the temperature parameter for each dataset.

\textbf{Our approach} PPT (Distribution), where fine-tuning was performed on full posterior probabilistic labels, is our primary method. 
We also consider a variant, called PPT (Mean), that fine-tunes on the same dataset but targets only the mean of the distribution. This ablates the distributional target and tests the utility of such a target. In addition, we implement \textit{forward sampling} to generate distributional targets, instead of full posterior inference, as it allows creation of orders-of-magnitude, bigger datasets due to its inherent scalability. 

\textbf{PPT transfers to external benchmarks} Table~\ref{tab:calibration} summarizes the results. 
PPT (Distribution) substantially reduces NLL relative to the base model across all reported benchmarks for both \textsc{Llama} and \textsc{Qwen}.
The calibration results show a stronger and more consistent pattern on MMLU, HellaSwag, ARC-Challenge, and Winogrande. 
ECE improves substantially: PPT reduces ECE relative to the base model on every listed multiple-choice benchmark. This suggests that probabilistic-program-mediated fine-tuning improves the model's raw uncertainty estimates. 

Temperature scaling (TS) remains a strong post-hoc calibration baseline. In several cases, Base+TS closes much of the NLL and ECE gap between the base and fine-tuned models. However, applying temperature scaling on top of PPT (Distribution) consistently yields further improvements, with lower NLL and ECE on most benchmarks. This indicates that PPT and post-hoc calibration are complementary: temperature scaling adjusts confidence on a target distribution, while PPT changes the underlying model to produce better raw uncertainty estimates.

Finally, PPT also improves accuracy on these benchmarks. In OpenEstimate, PPT (Distribution) substantially improves the prior elicitation accuracy: PPT (Distribution) reduces MAE from 29.0 to 23.3 for \textsc{Llama}. This is consistent with the goal of improving quantitative inductive estimation. Although the gains are modest for other benchmarks, our results suggest that improved prior knowledge elicitation and calibration can contribute to not only improving answer distribution, but also correctly flipping the argmax answer.

\textbf{Fine-tuning without distributional target reduces improvement} For both \textsc{Llama} and \textsc{Qwen}, PPT (Distribution) and PPT (Mean) use identical training data except for the target, but PPT (Distribution) outperforms on every metric. PPT (Distribution) also attains higher accuracy and stronger post-TS calibration than models fine-tuned with forward sampling data (Table \ref{tab:calibration_pos_fwd} in the Appendix). This suggests that distributional information plays an important role in the improvements observed in external benchmark transfers.

\section{Discussion}
\label{sec:disc}

LLMs have fared well in deductive reasoning tasks, such as mathematics and coding, but their performance on inductive reasoning, which requires inferring uncertain latent properties from sparse observations, has remained insufficient. While post-training offers the means to improve inductive inference, collecting high-quality data at scale and handling inherently distributional targets posed a significant challenge. To address these shortcomings, we propose program-based posterior training (PPT), a pipeline that (1) programmatically synthesizes diverse open-world scenarios; (2) translates these natural language scenarios to probabilistic programs;  (3) runs posterior inference on the generated program to produce distribution targets; and (4) fine-tunes an LLM on these data to approximate posterior beliefs.  
Overall, our results demonstrate that PPT allows models to internalize uncertainty, generalizing to entirely unseen motifs and domains, while remaining complementary to traditional temperature scaling techniques.     

More broadly, our results suggest a different way to think about reasoning post-training. Standard reasoning supervision is often framed around problems with single verifiable answers, but inductive reasoning requires models to maintain uncertainty over latent explanations and outcomes. 
PPT uses probabilistic programs to construct such a supervision.
Empirically, we found that our approach improves both task-specific inductive inference and broader uncertainty behavior. On held-out synthetic scenarios, PPT substantially reduces MAE relative to base LLMs and strong closed-source baselines, including settings with unseen motifs and a held-out healthcare domain. In human-labeled sports scenarios, PPT improves alignment with human judgments, reaching a correlation comparable to the original MSA programs. On external benchmarks, PPT improves OpenEstimate prior-elicitation MAE and consistently reduces NLL and ECE across multiple-choice tasks, with gains that remain complementary to temperature scaling. Together, these results suggest that probabilistic programs can serve not only as test-time reasoning tools, but also as scalable supervision sources for training LLMs to form calibrated beliefs.

\textbf{Limitations and future directions} 
Our approach relies on LLM-synthesized probabilistic programs, so the quality of the supervision depends on whether the generated programs faithfully capture the intended scenario. Although we selectively evaluated some of the generate programs against similar open-source programs and human results, future work can add stronger automatic validation and human audits of program quality.

Our experiments focus on numeric and multiple-choice queries, which allow precise measurement of posterior distributions, MAE, NLL, and ECE. This does not fully cover open-ended inductive reasoning. Extending PPT to more naturalistic settings with intermediate reasoning steps and/or interactions is an important direction.

\textbf{Conclusion}
Our results show that using LLMs to generate probabilistic programs that are in turn used to train LLMs on natural-language inductive inference problems is an effective way of improving the capabilities of these models. Fine-tuning on data generated from probabilistic programs improves held-out probabilistic estimation, alignment with human judgments, transfer to external estimation and multiple-choice benchmarks, and raw calibration. These results suggest that probabilistic programs can serve not only as test-time reasoning tools, but also as scalable supervision sources for training LLMs to form calibrated beliefs.


\section*{Acknowledgments}
LZ and TLG are supported by grant
N00014-23-1-2510 from the Office of Naval Research. AKJ is supported by a Natural and Artificial Mind (NAM) Fellowship from the Scully Peretsman foundation. BML is supported by the U.S. National Science Foundation (NSF) under Cooperative Agreement No.~2433429, NSF AI Research Institute on Interaction for Al Assistants (ARIA). We thank Katie Collins for helpful discussion.

\bibliographystyle{unsrt} 
\bibliography{references}

\newpage


\appendix

\section{Technical appendices and supplementary material}

\subsection{Implementational details}
\label{sec:appendix-imp}

\textbf{Hyperparameters}
Here we detail the experimental setup used in our experiments. We use PyTorch with Torchtune to fine-tune \textsc{Llama-3-8B-Instruct} and \textsc{Qwen-2-7B-Instruct}. Each experiment uses a single A100 GPU with 40GB memory. All methods use the AdamW optimizer \citep{loshchilov2017decoupled}, batch-size of 2, inner-loops with 5 gradient steps, LoRA adapters with rank $=8$ following the standard practice, and learning rate is tuned in $\small{[10^{-6}, 10^{-4}]}$. 

An `epoch' here is defined as training on 200 batches from the train set, and all models train up to 500 epochs, using an early-stopping of 50 epochs.

\textbf{Scenario examples}
Here we show a scenario and query example from each domain: sports, healthcare, general. While most scenarios follow a certain writing format, we also show one free-form scenario.

Here is an instruction block that the LLM sees when responding to each scenario.
\begin{mycallout}{LLM Instruction}

Answer the query in the scenario and return only an integer wrapped in \textless\ and \textgreater. For example,         
\textless x\textgreater. Use 0-100 scale. For a query on individual rank, a higher number means a higher ranking (e.g.\
100 means the individual ranks highest in that criterion; 1 is lowest). For a query on which of the two teams wins, a    
smaller number means the first team more likely wins.

\end{mycallout}

A sports scenario example is given in the main text in Figure \ref{fig:example-posterior}. 

Here are the other scenarios. We also add the program-inferred posterior mean as answers for illustration.
\begin{mycallout}{Healthcare Scenario}
BACKGROUND\\
In this endocrine day clinic, patients are monitored during a series of metabolic stress tests. In each episode,
the group that responds better depends on the average insulin sensitivity / glucose regulation efficiency and
autonomic stability (orthostatic tolerance) of the patients, modulated by missed meals / fasting that episode,
acute stress level that episode, and caffeine/alcohol intake that episode. Patients are under the care of either
Dr. Smith or Dr. Jones, and all clinical events take place on the same day.\\

CONDITIONS\\
Dr. Smith cares for Taylor, Ava, Beth, Cara, Dana, Elsa, Faye, Gina, Hope, Iris, Judy, Kara, Lana, and Mary,
whereas Dr. Jones cares for Nina, Opal, Page, Quin, Rosa, Sara, Tara, Uma, Vera, Willa, Xena, Yara, and Zara.\\
In the first episode, Taylor, Ava, Beth, Cara, Dana, Elsa, and Faye responded better than Gina, Hope, Iris, Judy,
Kara, Lana, and Mary.\\
In the second episode, Taylor, Nina, Opal, Page, Quin, Rosa, and Sara responded better than Tara, Uma, Vera,
Willa, Xena, Yara, and Zara.\\
In the third episode, Taylor, Gina, Hope, Iris, Judy, Kara, and Lana responded better than Nina, Opal, Page,
Quin, Rosa, Sara, and Tara.\\

QUERIES\\
Query 1: On a percentage scale from 0 to 100\%, how strong was missed meals / fasting that episode in the third
clinical episode for Gina?\\
Answer 1: 73\\
Query 2: In a new clinical episode later this same day involving Gina, Hope, Iris, Judy, Kara, Lana, and Mary
versus Nina, Opal, Page, Quin, Rosa, Sara, and Tara, who would have the better outcome and by how much?\\
Answer 2: 6\\
Query 3: In a new clinical episode later this same day involving Taylor, Ava, Beth, Cara, Dana, Elsa, and Faye
versus Tara, Uma, Vera, Willa, Xena, Yara, and Zara, who would have the better outcome and by how much?\\
Answer 3: 5\\
Query 4: Out of 100 random patients, where do you think Taylor ranks in terms of intrinsic insulin sensitivity /
glucose regulation efficiency?\\
Answer 4: 68
\end{mycallout}
\begin{mycallout}{General Domain Scenario}
BACKGROUND\\
In this event, watchmakers are competing in a series of timed challenges to assemble intricate mechanical
watches. In each round, the team that achieves the better outcome depends on the average assembly precision of
the watchmakers, based on their intrinsic mechanical dexterity modulated by four other factors: lighting quality,
tool sharpness, eye fatigue, and ambient humidity. Watchmakers compete in teams of five, and all challenges take
place on the same day.\\

CONDITIONS\\
In the first challenge, Julian, Clara, Thomas, Elise, and Victor completed the watch assembly more successfully
than Marcus, Sophie, Daniel, Chloe, and Felix.
In the second challenge, Liam, Nora, Oliver, Maya, and Lucas completed the watch assembly more successfully than
Henry, Emma, Wyatt, Grace, and Felix.
In the third challenge, Julian, Clara, Thomas, Elise, and Victor completed the watch assembly more successfully
than Liam, Nora, Oliver, Maya, and Felix.\\

QUERIES\\
Query 1: On a percentage scale from 0 to 100\%, how strong was eye fatigue in the third event for Felix?\\
Answer 1: 66\\
Query 2: In a new situation later this same day involving Liam, Nora, Oliver, Maya, and Lucas (Team 1) versus
Marcus, Sophie, Daniel, Chloe, and Felix (Team 2), who would have the better outcome and by how much?\\
Answer 2: 30\\
Query 3: In a new situation later this same day involving Julian, Clara, Thomas, Elise, and Victor (Team 1)
versus Henry, Emma, Wyatt, Grace, and Lucas (Team 2), who would have the better outcome and by how much?\\
Answer 3: 22\\
Query 4: Out of 100 random watchmakers, where do you think Felix ranks in terms of intrinsic mechanical
dexterity?\\
Answer 4: 34
\end{mycallout}
\begin{mycallout}{Free-Form Scenario}
\textbf{Seaside Pickleball Invitational: Coaching Staff Chat Log}

\textbf{Head Coach Martinez:} Let's review today's tournament results. How did our fixed doubles pairs look out
there?

\textbf{Asst Coach Davis:} It was an interesting day, definitely heavily influenced by the coastal weather. The
morning started dead calm and cool. Tariq and Leo played Jin and Carlos first. Tariq and Leo won comfortably.
Tariq’s baseline drives were incredibly sharp right out of the gate.

\textbf{Head Coach Martinez:} Good start. How about the mid-day matches?

\textbf{Asst Coach Davis:} The sun got absolutely brutal around noon. Tariq and Leo played Emma and Noah. Tariq
and Leo won narrowly. To be honest, Leo looked like he was really dragging his feet in the heat and missing easy
kitchen volleys that he normally puts away.

\textbf{Head Coach Martinez:} I actually caught the singles exhibition right after lunch. Tariq played Jin
1-on-1. Tariq won by a lot. He was covering the whole court and barely looked like he was breaking a sweat.

\textbf{Asst Coach Davis:} Yeah, individually, Tariq is a machine. But the late afternoon doubles match was a
different story. Tariq and Leo went up against Sarah and Maya. By then, the coastal wind was howling, which was
completely messing with the lightweight ball on lobs and drops. Plus, it was Tariq and Leo's third doubles match
of the day in the sun. Sarah and Maya ended up winning narrowly.

\textbf{Head Coach Martinez:} Tough break, but it makes sense. Sarah and Maya play a very grounded, tactical game
that probably holds up well when the wind picks up. Did any other matches happen before sunset?

\textbf{Asst Coach Davis:} Just one evening match, right after the wind completely died down and the temperature
dropped back to normal. Jin and Carlos played Emma and Noah. Jin and Carlos won comfortably, looking very
coordinated.\\

***

\textbf{Post-Tournament Analytical Assessment}

QUERIES\\
Query 1: What is the probability (from 0 to 100) that Tariq and Leo would comfortably defeat Sarah and Maya if
they played a rematch first thing the next morning in calm, cool conditions?\\
Answer 1: 0\\
Query 2: What is the probability (from 0 to 100) that Jin and Carlos would narrowly defeat Sarah and Maya if they
played a match against each other under standard, weather-neutral conditions?\\
Answer 2: 0\\
Query 3: Out of 100 random competitive pickleball players, where would you rank Tariq's intrinsic individual
skill?\\
Answer 3: 91\\
Query 4: On a scale from 0 to 100 (where 0 is completely fresh and 100 is completely exhausted), what was Leo's
likely fatigue level during the late afternoon match against Sarah and Maya?\\
Answer 4: 23
\end{mycallout}

\textbf{Prompts to LLM in data generation}
This section shows further implementational details based on Figure \ref{fig:msa}. Step 1 - scenario generation prompts are shown below. Our step 2 and step 3 prompts follow those used by \cite{wong2025modeling}.

\begin{mycallout}{Example Sports Scenario Used In-Context}
\textless{}START\_SCENARIO\textgreater{} \\
BACKGROUND \\
In this event, the athletes are competing in a series of synchronized diving tournaments. Each tournament consists of a series of rounds. In each match, athletes compete as part of a team. In a given round, each team receives a combined score based on the difficulty of the dive and the execution of the dive. Athletes compete either individually or as a team. \\
All matches take place on the same day.

CONDITIONS \\
In the first round, Jamie and Gale beat Sam and Jordan. \\
In the second round, Jamie and Gale beat Blake and Avery. \\
In the third round, Jamie and Avery beat Blake and Sam. \\
In the fourth round, Cameron and Gale beat Dakota and Joe. \\
In the fifth round, Jamie and Cameron lost to Blake and Avery.

QUERIES \\
Query 1: Out of 100 random athletes, where do you think Jamie ranks in terms of intrinsic skill? \\
Query 2: On a percentage scale from 0 to 100\% (where 0=extremely easy, 100=as difficult as possible), how difficult of a dive do you think the team Jamie was on in the third round attempted? \\
Query 3: In a new round later this same day between Jamie and Gale (Team 1) and Sam and Alice (Team 2), who would win and by how much? \\
Query 4: In a new round later this same day between Jamie and Cameron (Team 1) and Avery and Drew (Team 2), who would win and by how much? \\
\textless{}END\_SCENARIO\textgreater{}
\end{mycallout}

\begin{mycallout}{Sports Scenario Generation Prompt}
You will be asked to generate a scenario based on the given example scenario. A scenario consists of a background, 3 or 4 conditions, and 4 queries. Generate scenarios in the domain of sports. Keep the type of queries the same. In the conditions, keep them relatively concise similar to the example, and do not exactly specify the win and loss points. Also, prioritize the form ``A beat B'' compared to ``A beat B 1 on 1'' or ``A beat B in a singles match''. Notice that the background discusses that performance depends on individual-inherent variable(s), \textless{}MAIN\_VARIABLE\textgreater{}, and episodic variable(s), \textless{}EPISODIC\_VARIABLE\textgreater{}. You can improvise the phrasing, instead of saying ``performance depends on x and y''.

Create scenarios where, for one of the players denoted 'X', \textless{}P\textgreater{}, \textless{}C\textgreater{}, and \textless{}R\textgreater{}. Choose \textless{}N\textgreater{} for team-size. \textless{}1v1 OPTION\textgreater{}Use a random name for player 'X' (don't actually use 'X'). Query 1 should ask about the individual-inherent factor. Query 2 should ask about the episodic factor. 

Use the following sports subdomain for the scenario: \textless{}SUBDOMAINS\textgreater{}. Here is the example scenario, and be sure to use the same \textless{}START\_SCENARIO\textgreater{} \textless{}END\_SCENARIO\textgreater{} to wrap around the scenario:

<Example sports scenario here>
\end{mycallout}

\begin{mycallout}{Free-Form Scenario Generation Prompt}
You are generating a synthetic ``open-world'' sports scenario for training/evaluating probabilistic reasoning in LLMs. 

Core idea \\
- A scenario describes a small sports world where outcomes depend on (i) each athlete's intrinsic strength and (ii) multiple latent, match-dependent factors (e.g., effort, fatigue, teamwork chemistry, strategy matchups, weather, injuries, home advantage, equipment). \\
- Observations are incomplete and noisy: the reader must infer hidden traits and hidden match factors from a few results and contextual clues.

Freedom requirement (important) \\
- Do NOT use a rigid template like ``BACKGROUND / CONDITIONS / QUERIES''. \\
- You may vary format: a short story, a coach's notes, a sports journalist recap, a chat log, a referee report, a table of results, etc. \\
- Vary the number of observed events (e.g., 3--8 results) and the number of queries (e.g., 3--6). \\
- Keep the scenario self-contained.

Output formatting \\
- Produce exactly 1 scenario. \\
- Wrap the scenario in \textless{}START\_SCENARIO\textgreater{} ... \textless{}END\_SCENARIO\textgreater{}. \\
- Inside the scenario, include: \\
  (a) a narrative + observable match results, \\
  (b) enough detail to imply latent factors without explicitly revealing them, \\
  (c) a set of queries that require inference and uncertainty.

Sports world design constraints (must-have) \\
- Create a scenario where, for one of the players denoted 'X', \textless{}P\textgreater{}, \textless{}C\textgreater{}, and \textless{}R\textgreater{}. Use \textless{}N\textgreater{} for team-size. \textless{}1v1 OPTION\textgreater{}Use a random name for player 'X' (don't actually use 'X'). \\
- Choose from the following sports subdomains for this scenario: \textless{}SUBDOMAINS\textgreater{}. \\
- Do not specify exact point totals; if needed, use qualitative margins (``narrowly'', ``comfortably'', ``by a lot'', ``slightly'', ``in overtime'', ``came from behind''). \\
- You may include time ordering (``earlier that day'', ``two days later'') and let fatigue/learning matter.

Latent factors (must-have) \\
- The scenario must have at least TWO latent match-dependent factors beyond intrinsic strength. \\
- At least one factor should plausibly vary across matches (e.g. effort, fatigue, tilt, coordination, weather).

Queries (must-have) \\
Make queries that require: \\
1) Estimating an athlete's intrinsic strength as a distribution or percentile rank (e.g., ``out of 100 athletes, where do they rank?''). \\
2) Inferring a latent match factor on a 0--100 scale for a specific match (effort / teamwork / fatigue / etc.). \\
3) Predicting outcomes of 1--2 hypothetical future matches, including *who wins* and *by how much* qualitatively (or as a probability), while acknowledging uncertainty. \\
Important: each query should be answerable by one number. Wrong examples: 1) explicitly asking for verbal justification; 2) asking about two players' intrinsic strengths within one query. However, do not explicitly say 'answer the queries in one number'. It should be implicit in the queries like in the example scenario below.

Uncertainty instruction \\
- Ensure that multiple explanations remain plausible. Do not leak the true strengths or factor values. \\
- Avoid making any single athlete dominate all outcomes; include at least one ``surprising'' result that is explainable via latent factors.

An example scenario is given below. Remember to vary the form and story, but keep the query format the same.

<Example sports scenario here>

Now generate the scenario following all rules above.
\end{mycallout}

These are the \textbf{motifs} used:

P for probability / difficulty: motifs\_P = ['X consistently wins',
            'X consistently loses',
            'X wins all but one match',
            'X loses all but one match']

C for confounded teammates: motifs\_C = [
    'X always teams up with the same teammate(s)',
    'X teams up with different player(s) most of the times'
]

R for round-robin (team rotation): motifs\_R = [
    'players rotate across teams',
    'players have fixed teams',
    'players generally have fixed teams, except X'
]

\subsection{Additional experiments}
\label{sec:appendix-exp}

\begin{figure}
    \centering
    \includegraphics[width=0.6\linewidth]{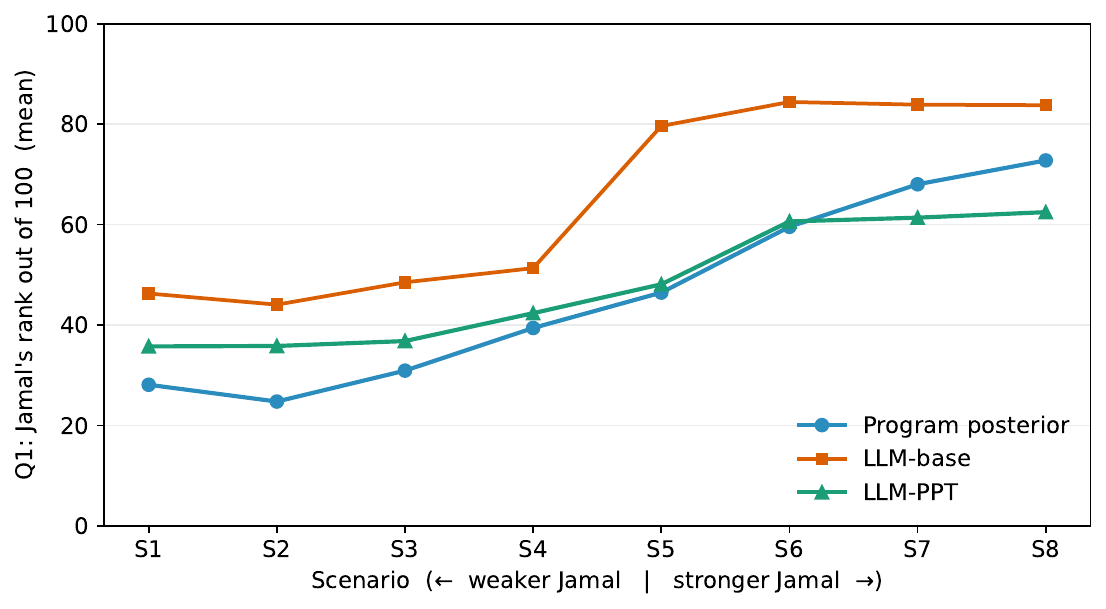}
    \caption{Mean estimates on Jamal's strength on eight scenarios that progressively goes from weaker to stronger Jamal. \textit{All methods in general correctly show a progressive increase in its mean estimate on Jamal's strength. However, base LLM plateaus more than the PPT-LLM, which still increases by 1-2 \% each step for the S7 and S8, as well as maintaining closer estimates with the program.}}
    \label{fig:progressive}
\end{figure}

\textbf{Progressively changing scenarios} We manually edit the scenario by using one action at a time (flip a win or loss or to swap two players), creating 8 scenarios S1 to S8 where player Jamal progressively goes from weaker to stronger. All methods in general correctly show a progressive increase in its mean estimate on Jamal's strength. However, base LLM plateaus more than the PPT-LLM, which still increases by 1-2 \% each step for the S7 and S8, as well as maintaining closer estimates with the program (Figure \ref{fig:progressive}).

\textbf{Posterior sampling and forward sampling} We compare two approaches we used to generate data for LLM fine-tuning: \textit{posterior sampling} and \textit{forward sampling}. Posterior sampling fine-tunes on a distributional target, whereas forward sampling, by construction, generates massive amounts of data with only a point estimate available for each query. 

Table \ref{tab:calibration_pos_fwd} shows that \textit{posterior sampling} attains better accuracy while \textit{forward sampling} attains comparable-to-better raw calibration and negative log-likelihood. Meanwhile, when temperature scaling (TS) is applied, \textit{posterior sampling} gains an edge on these two metrics.

\begin{table}[ht]
\centering
\caption{Transfer to external benchmarks on \textsc{Llama-3-8B}.
NLL/ECE include a temperature-scaling (TS) block.
Best per column in \textbf{bold} (within each block).}
\label{tab:calibration_pos_fwd}

\subcaption{NLL across benchmarks (lower is better).}
\label{tab:nll_pos_fwd}
\resizebox{\textwidth}{!}{%
\begin{tabular}{lllllllll}
\toprule
Method & OE & BT & BT-guided & MMLU & TruthfulQA & HellaSwag & ARC-C & Winogrande \\
\midrule
Posterior (Dist)     & \underline{\bm{$4.53$}}
                  \scriptsize{$\pm 0.02$}
                & $1.25$
                  \scriptsize{$\pm 0.01$}
                & $1.13$
                  \scriptsize{$\pm 0.00$}
                & $1.01$
                  \scriptsize{$\pm 0.01$}
                & $1.78$
                  \scriptsize{$\pm 0.02$}
                & \underline{\bm{$0.88$}}
                  \scriptsize{$\pm 0.02$}
                & \underline{\bm{$0.61$}}
                  \scriptsize{$\pm 0.01$}
                & \underline{\bm{$0.80$}}
                  \scriptsize{$\pm 0.01$} \\
Forward (Mean)  & $4.79$
                  \scriptsize{$\pm 0.01$}                 & \underline{\bm{$1.11$}}
                  \scriptsize{$\pm 0.01$}                 & \underline{\bm{$1.08$}}
                  \scriptsize{$\pm 0.00$}                 & \underline{\bm{$0.95$}}
                  \scriptsize{$\pm 0.00$}                 & \underline{\bm{$1.52$}}
                  \scriptsize{$\pm 0.03$}                 & $1.05$
                  \scriptsize{$\pm 0.01$}                 & $0.67$
                  \scriptsize{$\pm 0.01$}                 & $0.89$
                  \scriptsize{$\pm 0.02$} \\
\midrule
Posterior (Dist) + TS    & ---
                & \underline{\bm{$1.08$}}
                  \scriptsize{$\pm 0.00$}
                & \underline{\bm{$1.09$}}
                  \scriptsize{$\pm 0.00$}
                & \underline{\bm{$0.96$}}
                  \scriptsize{$\pm 0.01$}
                & \underline{\bm{$1.62$}}
                  \scriptsize{$\pm 0.03$}
                & \underline{\bm{$0.87$}}
                  \scriptsize{$\pm 0.02$}
                & \underline{\bm{$0.60$}}
                  \scriptsize{$\pm 0.01$}
                & \underline{\bm{$0.80$}}
                  \scriptsize{$\pm 0.00$} \\
Forward (Mean) + TS    & ---
                & $1.09$
                  \scriptsize{$\pm 0.00$}
                & \underline{\bm{$1.09$}}
                  \scriptsize{$\pm 0.00$}
                & \underline{\bm{$0.97$}}
                  \scriptsize{$\pm 0.00$}
                & $1.73$
                  \scriptsize{$\pm 0.05$}
                & $0.93$
                  \scriptsize{$\pm 0.01$}
                & $0.63$
                  \scriptsize{$\pm 0.01$}
                & \underline{\bm{$0.80$}}
                  \scriptsize{$\pm 0.00$} \\
\bottomrule
\end{tabular}%
}

\vspace{1em}

\subcaption{Expected Calibration Error (15-bin) (lower is better).}
\label{tab:ece_pos_fwd}
\resizebox{\textwidth}{!}{%
\begin{tabular}{llllllll}
\toprule
Method & BT & BT-guided & MMLU & TruthfulQA & HellaSwag & ARC-C & Winogrande \\
\midrule
Posterior (Dist)     & $0.218$
                  \scriptsize{$\pm 0.002$}
                & $0.134$
                  \scriptsize{$\pm 0.007$}
                & $0.112$
                  \scriptsize{$\pm 0.006$}
                & $0.275$
                  \scriptsize{$\pm 0.009$}
                & \underline{\bm{$0.064$}}
                  \scriptsize{$\pm 0.010$}
                & \underline{\bm{$0.058$}}
                  \scriptsize{$\pm 0.010$}
                & \underline{\bm{$0.094$}}
                  \scriptsize{$\pm 0.009$} \\
Forward (Mean)  & \underline{\bm{$0.091$}}
                  \scriptsize{$\pm 0.015$}                 & \underline{\bm{$0.040$}}
                  \scriptsize{$\pm 0.009$}                 & \underline{\bm{$0.043$}}
                  \scriptsize{$\pm 0.001$}                 & \underline{\bm{$0.105$}}
                  \scriptsize{$\pm 0.003$}                 & $0.222$
                  \scriptsize{$\pm 0.012$}                 & $0.126$
                  \scriptsize{$\pm 0.006$}                 & \underline{\bm{$0.072$}}
                  \scriptsize{$\pm 0.014$} \\
\midrule
Posterior (Dist) + TS    & \underline{\bm{$0.025$}}
                  \scriptsize{$\pm 0.002$}
                & \underline{\bm{$0.020$}}
                  \scriptsize{$\pm 0.003$}
                & $0.069$
                  \scriptsize{$\pm 0.004$}
                & \underline{\bm{$0.223$}}
                  \scriptsize{$\pm 0.014$}
                & \underline{\bm{$0.032$}}
                  \scriptsize{$\pm 0.003$}
                & \underline{\bm{$0.051$}}
                  \scriptsize{$\pm 0.010$}
                & \underline{\bm{$0.080$}}
                  \scriptsize{$\pm 0.002$} \\
Forward (Mean) + TS    & \underline{\bm{$0.027$}}
                  \scriptsize{$\pm 0.007$}
                & $0.038$
                  \scriptsize{$\pm 0.007$}
                & \underline{\bm{$0.045$}}
                  \scriptsize{$\pm 0.004$}
                & \underline{\bm{$0.206$}}
                  \scriptsize{$\pm 0.008$}
                & $0.069$
                  \scriptsize{$\pm 0.004$}
                & $0.066$
                  \scriptsize{$\pm 0.003$}
                & \underline{\bm{$0.082$}}
                  \scriptsize{$\pm 0.005$} \\
\bottomrule
\end{tabular}%
}

\subcaption{Accuracy across benchmarks; OpenEstimate (OE) reported as MAE (lower is better).}
\label{tab:accuracy_pos_fwd}
\resizebox{\textwidth}{!}{%
\begin{tabular}{lllllllll}
\toprule
Method & OE (MAE) $\downarrow$ & BT & BT-guided & MMLU & TruthfulQA & HellaSwag & ARC-C & Winogrande \\
\midrule
Posterior (Dist)     & \underline{\bm{$23.3$}}
                  \scriptsize{$\pm 0.3$}
                & \underline{\bm{$39.3\%$}}
                  \scriptsize{$\pm 0.0\%$}
                & $37.3\%$
                  \scriptsize{$\pm 0.2\%$}
                & \underline{\bm{$61.2\%$}}
                  \scriptsize{$\pm 0.3\%$}
                & \underline{\bm{$43.2\%$}}
                  \scriptsize{$\pm 0.7\%$}
                & \underline{\bm{$66.8\%$}}
                  \scriptsize{$\pm 1.2\%$}
                & \underline{\bm{$79.6\%$}}
                  \scriptsize{$\pm 0.5\%$}
                & \underline{\bm{$58.1\%$}}
                  \scriptsize{$\pm 0.4\%$} \\
Forward (Mean)  & $27.7$
                  \scriptsize{$\pm 0.0$}                 & $38.0\%$
                  \scriptsize{$\pm 0.5\%$}                 & \underline{\bm{$38.3\%$}}
                  \scriptsize{$\pm 0.6\%$}                 & $60.2\%$
                  \scriptsize{$\pm 0.2\%$}                 & \underline{\bm{$42.5\%$}}
                  \scriptsize{$\pm 0.7\%$}                 & $64.4\%$
                  \scriptsize{$\pm 0.6\%$}                 & $77.3\%$
                  \scriptsize{$\pm 0.4\%$}                 & $55.4\%$
                  \scriptsize{$\pm 0.3\%$} \\
\bottomrule
\end{tabular}%
}

\vspace{1em}
\end{table}


\end{document}